\documentclass[runningheads]{llncs}
\usepackage{fourier} 

\usepackage{makecell}

\usepackage{array}
\usepackage{colortbl}
\usepackage{graphicx}
\usepackage{amsmath,amssymb,amsfonts}
\usepackage{multirow}
\usepackage{colortbl}
\usepackage{algorithmic}
\usepackage{graphicx}
\usepackage{textcomp}
\usepackage{xcolor}
\newcolumntype{P}[1]{>{\centering\arraybackslash}p{#1}}
\DeclareMathOperator*{\argmin}{arg\,min}
\def\BibTeX{{\rm B\kern-.05em{\sc i\kern-.025em b}\kern-.08em
    T\kern-.1667em\lower.7ex\hbox{E}\kern-.125emX}}
\usepackage{amssymb}
\def\nbR{\ensuremath{\mathrm{I\! R}}}
\usepackage{cite}
\begin{document}
\title{Dairy Cow rumination detection: A deep learning approach}
%
%
\author{Safa Ayadi\inst{1,2} \and
Ahmed ben said\inst{1,3} \and
Rateb Jabbar \inst{1,3}\and 
Chafik Aloulou\inst{2} \and
Achraf Chabbouh\inst{1}  \and
Ahmed Ben Achballah \inst{1}}
\authorrunning{A. Safa et al.}
\institute{\scriptsize{LifeEye LLC, Tunis, Tunisia}
\email{\scriptsize{safa.ayadi,ahmedbs,rateb,achraf,ahmed@lifeye.io/}}\\
\url{\scriptsize{https://www.lifeye.io/}}
\and
\scriptsize{Faculty of Economics and Management of Sfax, University of Sfax, 3018 Sfax, Tunisia}
\email{\scriptsize{chafik.aloulou@fsegs.rnu.tn}}\\
\and
\scriptsize{Department of Computer Science \& Engineering, Qatar University, Doha, Qatar}\\
\email{\{abensaid,rateb.jabbar\}@qu.edu.qa}}
\maketitle              
\begin{abstract}
Cattle activity is an essential index for monitoring health and welfare of the ruminants. Thus, changes in the livestock behavior are a critical indicator for early detection and prevention of several diseases. Rumination behavior is a significant variable for tracking the development and yield of animal husbandry. Therefore, various monitoring methods and measurement equipment have been used to assess cattle behavior. However, these modern attached devices are invasive, stressful and uncomfortable for the cattle and can influence negatively the welfare and diurnal behavior of the animal. Multiple research efforts addressed the problem of rumination detection by adopting new methods by relying on visual features. However, they only use few postures of the dairy cow to recognize the rumination or feeding behavior. In this study, we introduce an innovative monitoring method using Convolution Neural Network (CNN)-based deep learning models. The classification process is conducted under two main labels: ruminating and other, using all cow postures captured by the monitoring camera. Our proposed system is simple and easy-to-use which is able to capture long-term dynamics using a compacted representation of a video in a single 2D image. This method proved efficiency in recognizing the rumination behavior with 95\%, 98\% and 98\% of average accuracy, recall and precision, respectively.

\end{abstract}
%
%
\section{Introduction}
Cattle products are among the most consumed products worldwide (i.e., meat and milk) ~\cite{bouwman2005exploring}, which makes dairy farmers pressured by the intensity of commercial farming demands to optimize the operational efficiency of the yield system. Therefore, maintaining cattle physiological status is an important task to maintain an optimal milk production. It is widely known that rumination behavior is a key indicator for monitoring health and welfare of ruminants ~\cite{thomsen2004negative ,stangaferro2016use}. When the cow is stressed ~\cite{vandevala2017psychological}, anxious ~\cite{nolen2000role}, suffering from severe disease or influenced by any several factors, including the nutrition diet program ~\cite{grinter2019validation ,suzuki2014effect}, the rumination time will decrease accordingly. Early detection of any abnormality will prevent severe outcomes of the lactation program. Furthermore, the saliva produced while masticating aids to improve the rumen state ~\cite{beauchemin1991ingestion}. Rumination time helps farmers to predict estrus ~\cite{reith2014simultaneous} and calving ~\cite{paudyal2016peripartal,calamari2014rumination} period of dairy cows. It was proved that the rumination time reduces on the $14^{th}$ and $7^{th}$ day before calving ~\cite{paudyal2016peripartal} and decreases slowly three days before estrus \cite{reith2014simultaneous}. By predicting calving moments, the farmer/veterinarian would be able to maintain the health condition of the cow and prevent risks of any disease (e.g., Calf pneumonia) that could be mortal when the cow is having a difficult calving \cite{calamari2014rumination}. 
\newline
In previous decades, farmers performed a direct observation to monitor rumination \cite{krause1998fibrolytic}. However, this method has shown many limitations; it is time-consuming and requires labor wages, especially on large-sized farms. In modern farms, many devices based on sensors have been used to automatically monitor animal behavior such as sound sensors \cite{lopreiato2020post}, noseband pressure sensors \cite{shen2020rumination} and motion sensors \cite{mao2019automatic,shen2019automatic}. However, many of these sensors are designed to extract only a few behavioral patterns (e.g., sound waves), which developed the need of devising an automated system as a mean to assess health and welfare of animals and reduce operational costs for farmers. Machine Learning is able to extract and learn automatically from large-scale data using for example sophisticated Neural Networks (NNs). NNs are mainly used in Deep Learning algorithms that handily become state-of-the-art across a range of difficult problem domains \cite{jabbar2020driver,alhazbi2020using,said2018deep,abdelhediprediction}. Thus, the use of these developed technologies can improve the monitoring process and achieve an efficient performance in recognizing animal behavior. One of the most common used type of Deep Neural Networks for visual motion recognition is the convolutional neural networks (CNNs). CNNs can automatically recognize deep features from images and accurately perform computer vision tasks \cite{chen2018domain,zhang2019two}. Aside from these continuous achievements, these technologies require further improvement due to their lack of precision.
\newline
This work proposes a monitoring method to recognize cow rumination behavior. We show that CNN can accurately perform an excellent classification performance using an easy-to-use extension of state-of-the-art base architectures. Our contributions are as follow:
\begin{itemize}
\item We propose a simple and easy-to-use method that can capture long-term dynamics through a standard 2D image using dynamic images method \cite{bilen2016dynamic}.
\item With a standard deep learning CNN-based model, we accurately performed the classification tasks using all postures of the cows. 
\item We conduct comprehensive comparative study to validate the effectiveness of the proposed methodology for cow rumination detection. 
\end{itemize}
$\newline$
The remainder of this paper is organized as follows. The related works of this study are presented in Section 2. The developed method and the used equipment are described in detail in Section 3. The implementation of the model, the evaluation of the yielded results and the comparison with the state-of-the-art are discussed in Section 4. Finally, a conclusion and directions for future research are presented in Section 5.

\section{Related works}
In this section, we review existing research works, equipment and methods that addressed the challenging problem of rumination detection. The existing intelligent monitoring equipment can be split into four categories.

\subsection{Sound sensor for rumination detection}
The monitoring method with sound sensor, is mainly used to identify the rumination behavior by planting a microphone around the throat, forehead or other parts of the ruminant to record chewing, swallowing or regurgitating behavior. In fact, acoustic methods exhibit excellent performance in recognizing ingestive events. Milone et al.\cite{milone2012automatic} created an automated system to identify ingestive events based on hidden Markov models. The classification of chew and bite had an accuracy of 89\% and 58\% respectively. Chelotti et al. \cite{chelotti2018pattern} proposed a Chew-bit Intelligent Algorithm (CBIA) using sound sensor and six machine learning algorithms to identify three jaw movement events. This classification achieved 90\% recognition performance using the Multi-Layer Perceptron. Clapham et al. \cite{clapham2011acoustic} used manual identification and sound metrics to identify the jaw movement that detected 95\% of behavior, however this system requires manual calibration periodically which is not recommended for automated learning systems. Furthermore, some systems use sound sensors to recognize the rumination and grazing behavior after analysing jaw movement \cite{chelotti2020online,rau2020developments}. The monitoring methods with sound sensor gave a good performance. However, owing to their high-cost and trends in the distributed signals that negatively affect event detection, these devices are primarily used for research purposes.

\subsection{Noseband pressure sensor for rumination detection}
The monitoring method with a noseband pressure sensor, generally used to recognize rumination and feeding behavior using a halter and a data logger to record mastication through picks of pressure. Shen et al. \cite{shen2020rumination} used noseband pressure as core device to monitor the number of ruminations, the duration of rumination and the number of regurgitated bolus and achieved 100\%, 94,2\% and 94.45\% respectively as results of recognition. Zehner et al. \cite{zehner2017system} created two software to classify and identify the rumination and eating behavior using two versions of RumiWatch \footnote{https://www.rumiwatch.ch/} noseband pressure sensors. The achieved detection accuracy 96\% and 91\% of rumination time and 86\% and 96\% of feeding time for 1h resolution data provided by two noseband devices. The obtained results are important with these technologies however; the monitoring process is only useful for short-term monitoring and requires improvements to efficiently monitor the health and the welfare of animals.

\subsection{Triaxial acceleration sensor for rumination detection}
The monitoring method with a triaxial acceleration sensor that can, recognize broader sets of movement at various scales of rotation. It is common to use accelerometer sensors for its low cost. Shen et al. \cite{shen2019automatic} used triaxial acceleration to collect jaw movement and classify them into three categories: feeding, ruminating and other using three machine learning algorithms. Among them, the K-Nearest Neighbour (KNN) algorithm scored the best performance with 93.7\% of precision. Another work focused on identifying different activities of the cow using Multi-class SVM and the overall model performed 78\% of precision and 69\% of kappa \cite{martiskainen2009cow}. Rayas-Amor et al. \cite{rayas2017triaxial} used the HOBO-Pendant G-three-axis data recorder to monitor grazing and rumination behavior. The system recognized 96\% and 94.5\% respectively of 20 variances in visual observation per cow/day. The motion-sensitive bolus sensor was applied by Andrew et al. \cite{hamilton2019identification} to measure jaw motion through the bolus movement using SVM algorithm. This algorithm managed to recognize 86\% of motion. According to these findings, the accelerometer made an interesting performance in recognizing behavior; however, it still confuses activities of animals that share the same postures.

\subsection{Video equipment sensor for rumination detection}
The monitoring method with video equipment, recognize ruminant behavior by recording cow movement and extracting visual motions to identify and classify the animal behavior. According to the state-of-the-art, many initiatives focused on detecting the mouth movement using the optical flow technique that can detect motion from two consecutive frames. Mao et al. \cite{mao2019automatic} used this technique to track the rumination behavior of dairy cows automatically. This method reached 87.80\% of accuracy. Another work by Li et al. \cite{li2019tracking} on tracking multiple targets of cows to detect their mouth areas using optical flow technique achieved 89.12\% of the tracking rate. The mean shift \cite{cheng1995mean} and STC algorithms \cite{zhang2014fast} were used by Chen et al. \cite{yujuan2017intelligent,chen2018automatic} to monitor the rumination time using the optical flow technique to track the mouth movement of the cow. The monitoring process achieved 92.03\% and 85.45\% of accuracy, respectively. However, the learning process of these two methods is based only on the prone position and thus, it is not possible to monitor the diurnal rumination behavior in its different pastures. On other hand, the mouth tracking method can easily be influenced by cow movement which creates inferences in the training stage. CNN is another technique to extract features from images without any manual extraction. This technology is generally used for object detection \cite{achour2020image} and visual action recognition \cite{chen2018automatic,li2019mounting}. D Li et al. \cite{li2019mounting} used KELM \cite{huang2004extreme} to identify mounting behavior of pigs. This network achieved approximately 94.5\% of accuracy. Yang et al. \cite{yang2018feeding} applied Faster R-CNN \cite{ren2015faster} to recognize feeding behavior of group-housed pigs. The algorithm achieved 99.6\% of precision and 86.93\% of recall. Another recent work based on CNN was proposed by Belkadi at al. \cite{achour2020image}. It was developed on commercial dairy cows to recognize the feeding behavior, feeding place and food type. They implemented four CNN-based models to: detect the dairy cow, check availability of the food in the feeder and identify the food category, recognize the feeding behavior and identify each cow. This system is able to detect 92\%, 100\% and 97\% of the feeding state, food type and cow identity, respectively. Although the achieved performance is significant, this method is not suitable for detecting other behaviors since their used images focus only on the feeder area which boosted their performance. Overall, many of these proposed methods worked only with few postures of dairy cows to recognize the rumination or feeding behaviors. Conversely, video analysis methods can easily be influenced by weather conditions and other external, factors which causes noisy effects for the learning performance. These methods are more applicable to monitor cows housed indoors or for commercial purposes \cite{ambriz2015comparison}.

\subsection{Evaluation}
All the four categories performed well when it comes to recognizing animal behavior. However, many of these wearable devices are invasive, stressful and, accordingly, can influence the diurnal behavior of animals \cite{fenner2016effect}. Thus, using video equipment is more reliable and less invasive. In this work, we propose a method that relies on a non-stressful device and use a deep learning CNN-based method to recognize the rumination behavior of indoor-housed cows automatically. 
\begin{figure}
\includegraphics[width=\textwidth]{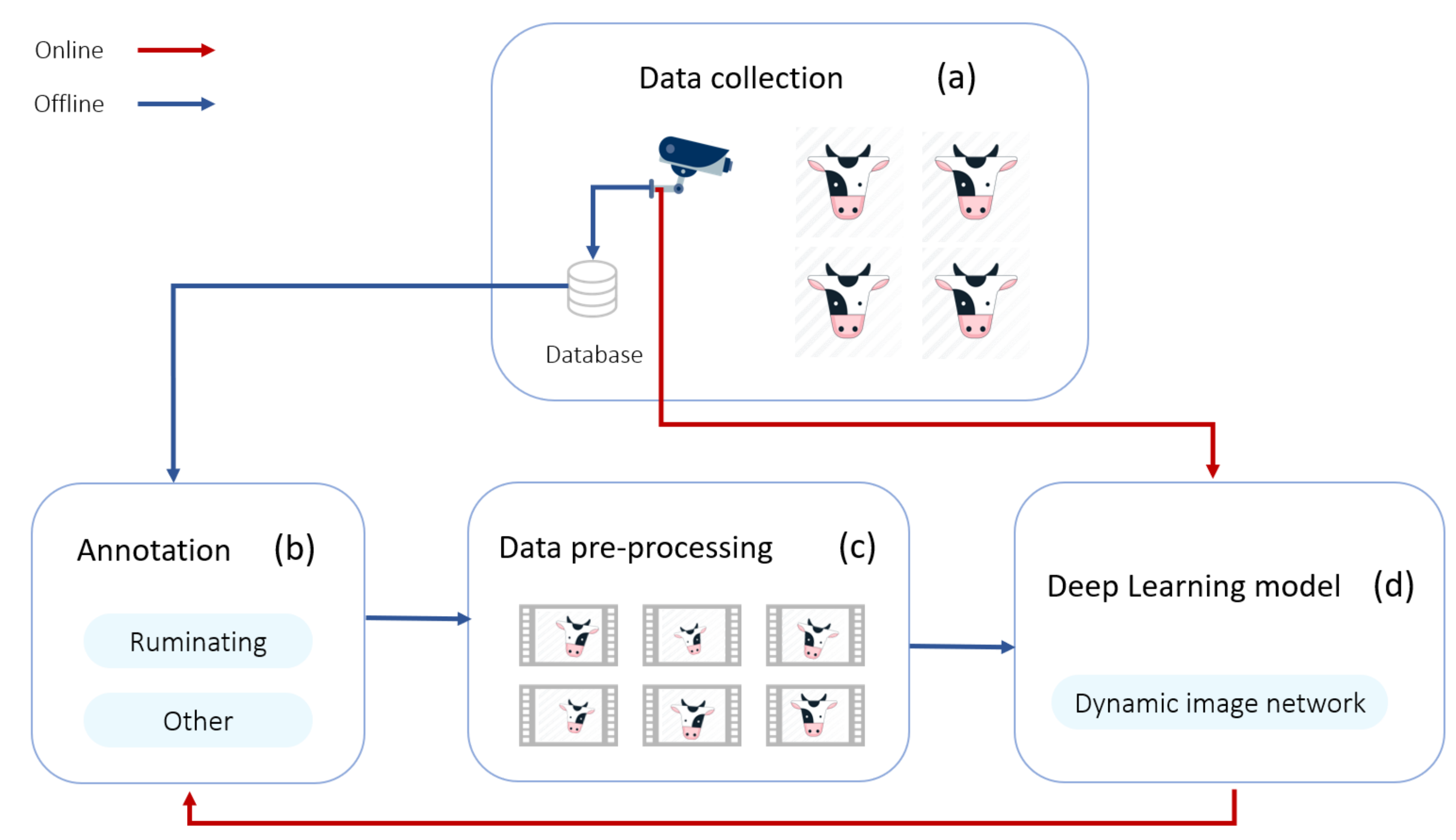}
\caption{The proposed system for cow rumination behavior recognition.}

\label{Proposedsystem}
\end{figure}
\section{Method and materials}
Our proposed system, is mainly constructed with four stages as depicted in Fig. \ref{Proposedsystem}. We use video equipment as a core device to collect cattle activities. The recorded videos, are continuously stored in the database and automatically segmented into frames. We collect data and carefully annotate it under two main labels (Section 3.1). Subsequently, these frames are cleaned from noisy effects and significantly generated to obtain a compacted representation of a video (Section 3.2). We apply the dynamic images approach that uses a standard 2D image as input to recognize dairy cow behavior (Section 3.3). This method can use a standard CNN-based architecture. All these processes were conducted offline. To implement and test our model, we choose several key architectures (Section 3.4) that gave relevant results in the classification stage. To avoid the overfitting of the model, we implemented a few regularization methods that can boost the performance of the network (Section 3.5).
\subsection{Data acquisition}
The research and experiments were conducted at the Lifeye LLC company\footnote{https://www.crunchbase.com/organization/lifeye} for its project entitled Moome\footnote{https://www.moome.io}, based in Tunis (Tunisia).The experimental subjects are Holstein dairy cows farmed indoor originating from different farms of rural areas from the south of Tunisia. Cattles were monitored by cameras planted in a corner offering a complete view of the cow body. The recorded videos were stored in an SD card, then, they were manually fed in the database storage and automatically divided into frames and displayed in the platform.
To ensure a real-time monitoring, cameras are directly connected to the developed system. The collected data includes 25,400 frames collected during the daytime and the unlit time in July 2019 and February 2020, then they were accurately distributed into two main labels according to each data folder content. Each data folder contains approximately 243 and 233 frames for 1 min video with a resolution of 640 × 480 pixels. Fig. \ref{fig2} illustrates examples of used frames. In fact, all captured cow postures were used for the training and testing sets, including, eating, standing, sitting, drinking, ruminating, lifting head and other movements. The definition of dairy cow rumination is presented in Table \ref{table:Labels}.

\begin{table}[]
	\centering
	\def\arraystretch{1.5}
	\caption{Definition of dairy cow rumination labels.}
	\label{table:Labels}
\begin{tabular}{|>{\hspace{0.5pc}}l|l<{\hspace{0.5pc}}|}
\hline
\textbf{Behavior}  & \textbf{Definition}                                                                   \\ \hline
Ruminating & The cow is masticating or swallowing of ingesta while sitting or standing.  \\ \hline
Other      & The cow is standing,   eating, drinking, sitting or doing any other activity. \\ \hline
\end{tabular}\quad
\end{table}

\begin{figure}
\includegraphics[width=\textwidth]{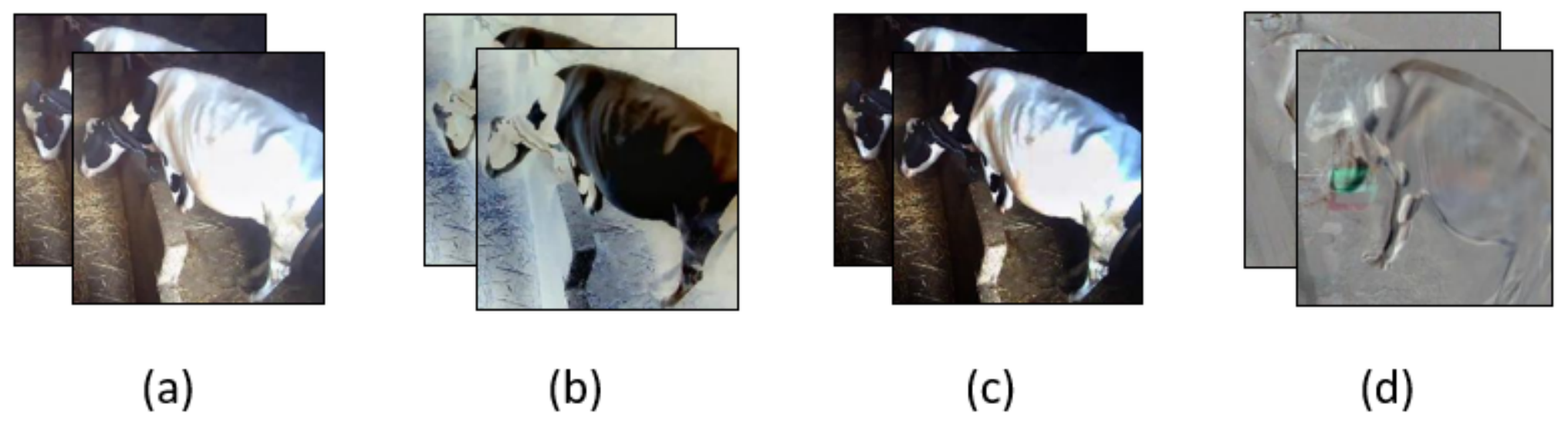}
\caption{Example of resultant frames from the pre-processing stage.}

\label{fig3}
\end{figure}

\subsection{Data pre-processing}
The aim of data pre-processing is to improve quality the frames by enhancing important image features or suppressing unwanted distortions from the image. In this study, the methods used for image pre-processing (Fig. \ref{Proposedsystem}c) including cropping, resizing, adding noises, data augmentation, and applying the dynamic image summarization method. The aim of cropping is delimiting the cow area by eliminating noisy pixels coming from sunlight or any noisy effects. Next, these cropped images were resized to 224 × 224 pixels (Fig.\ref{fig3}a) for the network training process. To ensure a good performance of the CNN model and test its stability, we added some noisy effects on images by randomly changing the brightness of images. In addition, to avoid overfitting issues, we applied the data augmentation technique by lightening the edges of the frames using negative effect (Fig. \ref{fig3}b) and gamma correction effect with 0.5 adjustment parameter (Fig. 3c). These corrections can be made even on low-quality images which can brighten the object threshold and facilitate the learning process. The obtained frames are generated using the dynamic image method. This method is able to summarize video content in single RGB image representation, using the rank pooling method \cite{bilen2016dynamic} to construct a vector $d^*$ that contains enough information to rank all T frames $\mathit{I}_{1},…,\mathit{I}_{T}$ in the video and make a standard RGB image (Fig. \ref{fig3}d) using the $RankSVM$ \cite{smola2004tutorial} formulation:

\begin{equation}
\begin{split}
d^*= p(\mathit{I}_{1},…,\mathit{I}_{T} ; \psi)= \argmin_{d}{E(d)} \\
E(d)=\frac{\lambda}{2}||d||^2+\frac{2}{T(T-1)}\sum_{q>t} \max{\{0,1-S(q|d)+S(t|d)\}}.
\end{split}
\end{equation}

$\newline$
Where $d$ $\in$ $ \nbR^d$ and $\psi(\mathit{I}_{t})$ $\in$ $ \nbR^d$ are vectors of parameters and image features, respectively while $\lambda$ is a regularization parameter. Up to time $t$, the time average of these features is calculated using $\mathit{V}_{t}\! = \frac{1}{t} \sum_{T=1}^{t}\psi(\mathit{I}_{T})$. The ranking function associates to each time $t$ a score $S(t|d)= \langle d,\mathit{V}_{t} \rangle$.The second term is constructed to test how many pairs are correctly ranked: if at least a unit margin is present, then the pair is well ranked, i.e. $S(q|d) > S(t|d)+1$ with $q>t$.

\begin{figure}
\centering
\includegraphics[width=110mm,scale=0.5]{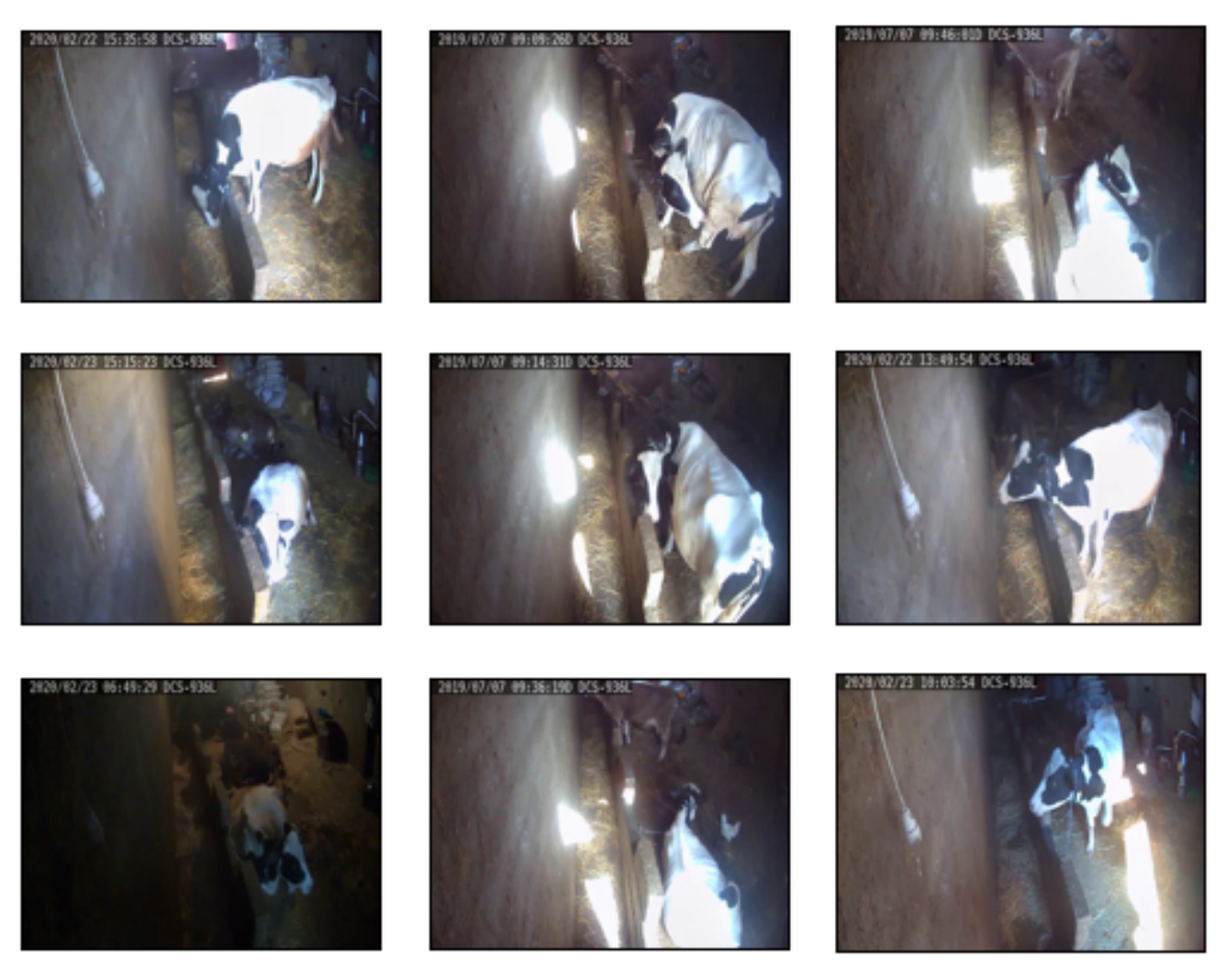}
\caption{Sample frames from the collected dataset.}

\label{fig2}
\end{figure}

\subsection{Dynamic image approach}
The dynamic image (Fig.\ref{Proposedsystem}d) is a CNN-based approach which powerfully recognizes motion and temporal features from a standard RGB image. It uses a compact representation of video that summarizes the motion of moving actors in a single frame. Interestingly, the dynamic image approach uses a standard CNN architecture pre-trained in still image Benchmark. This approach proved \cite{bilen2016dynamic} its efficiency in learning long-term dynamics and accurately performed 89.1\% of accuracy using the CaffeNet model trained on ImageNet and fine-tuned on UCF101 dataset \cite{soomro2012ucf101}.

\subsection{Key architectures}
To recognize rumination behavior of dairy cow, we used an end-to-end architecture that can efficiently recognize long-term dynamics and temporal features with a standard CNN architecture as it was presented in Section 3.3. To ensure good performance of our system, we chose to use only two well-known key architectures: VGG \cite{simonyan2014very} and ResNet \cite{he2016deep,he2016identity} that were adopted and tested in section 4. These two models are powerful and useful for image classification tasks. They achieved remarkable performance on ImageNet Benchmark \cite{krizhevsky2012imagenet} which make them the core of multiple novel CNN-based approaches \cite{donahue2015long,wu2020comprehensive}. The VGG model presents two main versions: VGG16 model with 16-layers and VGG19 model with 19-layers. ResNet model presents more than two versions that can handle a large number of layers with a strong performance using the so-called technique “identity shortcut connection” that enables the network to skip one or more layers.

\subsection{Overfitting prevention method}
Overfitting occurs when the model learns noises from the dataset while training, which make the learning performance much better on the training set than on the test set. To prevent these inferences, we adopted few regularization methods to improve the performance of the model. The first technique adopted is the dropout method \cite{srivastava2014dropout}, which can reduce interdependency among neurons by randomly dropping layers and connections during the training phase and thus forcing nodes within a layer to be more active and more adapted to correct mistakes from prior layers. The second technique is the data augmentation method, which prevents the model from overfitting all samples by increasing the diversity of images available for the training phase using different filters such as those presented in Section 3.3. The third technique is the early stopping method \cite{prechelt1998early}, which tracks and optimize the performance of the model by planting a trigger that stops the training process when the test error starts to increase and the train error starts decrease.

\begin{figure}
\includegraphics[width=\textwidth]{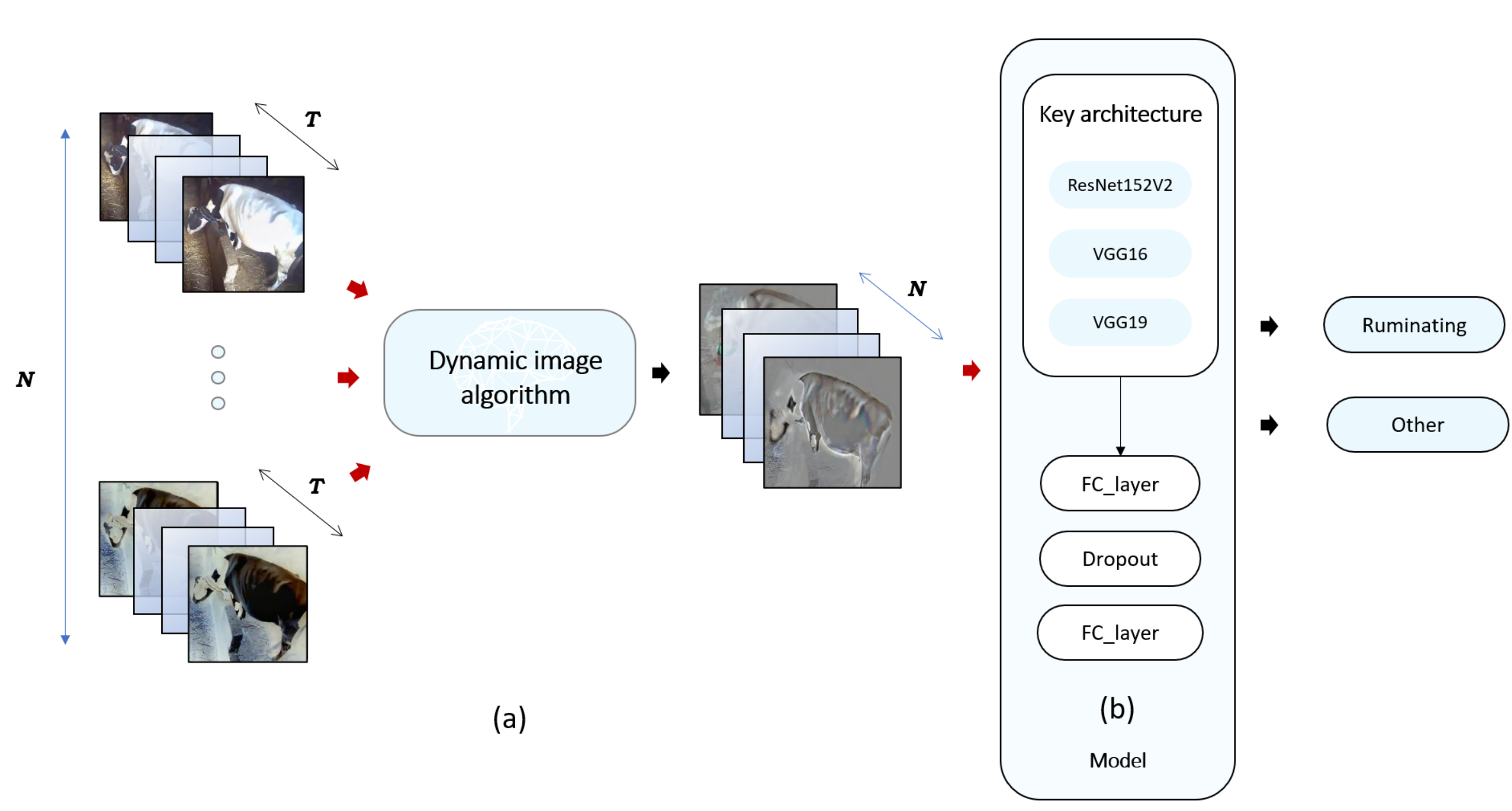}
\caption{Cow rumination behavior recognition procedures.}

\label{fig4}
\end{figure}

\section{Experiments}
In this section, we present the implementation process of the proposed model and the adopted evaluation metrics (Section 4.1). Subsequently, we evaluate the obtained results of rumination behavior recognition (Section 4.2). Finally, we compare the proposed model with other architectures (Section 4.3).

\subsection{Implementation}
We empirically evaluated the cow rumination behavior recognition using cow generated dataset as detailed in Section 3.2. For classification tasks, we implemented three pretrained CNN-base models: VGG16, VGG19, and ResNet152V2 to evaluate each model performance on the generated dataset. In the fine-tuning stage, we froze parameters of upper layers and replaced the rest of the layers by other layers as depicted in Fig. $\ref{fig4}$b. The dropout ratio was set to 0.5 to prevent overfitting. We used Adam optimizer \cite{kingma2014adam} with an intial learning rate $\mathit{lr}_{0}=0.001$ and eventually change its value during the training stage using the exponential decay formula:
\begin{equation} \label{lrr}
lr= {\mathit{lr}_{0}}\times{e^{kt}}
\end{equation}

Where t and k correspond to the iteration number and the decay steps, respectively. Models were trained on GPUs with batch size=12.
\newline
Let  $T = \{25,50,100\}$ be the number of frames used to generate a dynamic image (Fig. \ref{fig4}a). The aim is to evaluate the performance of the network with short video sequences. As for the rest of this study, we refer the datasets that contains dynamic images generated from 25, 50, and 100 frames for a single image as T25, T50 and T100, respectively. 

\subsection{Evaluation approach}
In Table \ref{table:Resultscowrumination}, the evaluation stage is made of two trials: in trial 1, we tested the model only on the generated data without data augmentation. In trial 2, we added more generated frames using the data augmentation technique. The whole generated data were divided into training and testing sets. In each trial, we evaluated the model performance based on accuracy, validation accuracy (val\_acc), loss and validation loss (val\_loss) results as metrics to measure model efficiency. Then, we evaluated the precision, the sensitivity and AUC metrics. The accuracy is one of the most common used metrics that count the percentage of correct classifications for the test data and it is calculated using Eq. (\ref{ac}). The loss value calculates the error of the model during the optimization process. The precision metric is obtained by Eq. (\ref{pr}) is consistent with the percentage of the outcomes. The sensitivity stands for the percentage of the total relevant results that are correctly classified. It is expressed using Eq. (\ref{sen}). The Area Under Curve AUC reflects how much the model is capable to distinguish between classes. The higher the AUC, the better the network is predicting classes. To measure the effectiveness of the model, machine learning uses the confusion matrix which contains four main variable: True Positive (TP), True Negative (TN), False Positive (FP) and False Negative (FN).

\begin{equation} \label{ac}
  Accuracy=\frac{TP + TN}{TP + FP + TN + FN}
\end{equation}

\begin{equation} \label{pr}
 Precision=\frac{TP}{TP + FP}
\end{equation}

\begin{equation} \label{sen}
 Sensitivity=\frac{TP}{TP + FN}
\end{equation}

\subsection{Evaluation results}

\begin{table}
\centering
\def\arraystretch{1.5}
\arrayrulecolor{black}
\caption{Results of cow rumination behavior model.}
\label{table:Resultscowrumination}
\begin{tabular}{!{\color{black}\vrule}l!{\color{black}\vrule}l!{\color{black}\vrule}l!{\color{black}\vrule}l!{\color{black}\vrule}l!{\color{black}\vrule}l!{\color{black}\vrule}l!{\color{black}\vrule}l!{\color{black}\vrule}} 
\hline
\textbf{Trial}              & \textbf{N° frames}              & \textbf{Key architecture} & \textbf{Dataset size}                       & \textbf{loss}   & \textbf{Val\_loss} & \textbf{Accuracy} & \textbf{Val\_acc}  \\ 
\hline
\multirow{9}{*}{1} & \multirow{3}{*}{T=25}  & ResNet152V2      & \multirow{3}{*}{\thead{N=1015 \\
  Test=213}} & 0.0359 & 0.7463    & 98.73\%  & 84.04\%   \\ 
\cline{3-3}\cline{5-8}
                   &                        & VGG-16           &                                    & 0.2081 & 0.2922    & 91.34\%  & 90.61\%   \\ 
\cline{3-3}\cline{5-8}
                   &                        & VGG-19           &                                    & 0.2874 & 0.3241    & 88.17\%  & 88.73\%   \\ 
\cline{2-8}
                   & \multirow{3}{*}{T=50}  & ResNet152V2      & \multirow{3}{*}{\thead{N=508 \\
  Test=107}}  & 0.0207 & 0.7929    & 99.86\%  & 82.24\%   \\ 
\cline{3-3}\cline{5-8}
                   &                        & VGG-16           &                                    & 0.1697 & 0.3679    & 92.39\%  & 85.98\%   \\ 
\cline{3-3}\cline{5-8}
                   &                        & VGG-19           &                                    & 0.2453 & 0.3600    & 88.45\%  & 86.92\%   \\ 
\cline{2-8}
                   & \multirow{3}{*}{T=100} & ResNet152V2      & \multirow{3}{*}{\thead{N=254\\
  Test=53}}   & 0.0050 & 0.7851    & 100\%    & 84.91\%   \\ 
\cline{3-3}\cline{5-8}
                   &                        & VGG-16           &                                    & 0.1200 & 0.4572    & 95.76\%  & 86.79\%   \\ 
\cline{3-3}\cline{5-8}
                   &                        & VGG-19           &                                    & 0.1363 & 0.3805    & 95.20\%  & 84.91\%   \\ 
\hline
\multirow{9}{*}{2} & \multirow{3}{*}{T=25}  & ResNet152V2      & \multirow{3}{*}{\thead{N=2030\\
  Test=426}} & 0.1153 & 0.8742    & 95.46\%  & 81.46\%   \\ 
\cline{3-3}\cline{5-8}
                   &                        & VGG-16           &                                    & 0.1370 & 0.3706    & 94.19\%  & 90.85\%   \\ 
\cline{3-3}\cline{5-8}
                   &                        & VGG-19           &                                    & 0.2186 & 0.3045    & 90.78\%  & 88.97\%   \\ 
\cline{2-8}
                   & \multirow{3}{*}{T=50}  & ResNet152V2      & \multirow{3}{*}{\thead{N=2032 \\
  Test=427}} & 0.0449 & 0.3687    & 98.12\%  & 88.13\%   \\ 
\cline{3-3}\cline{5-8}
                   &                        & VGG-16           &                                    & 0.0794 & 0.1944    & 96.91\%  & 93.91\%   \\ 
\cline{3-3}\cline{5-8}
                   &                        & VGG-19           &                                    & 0.1375 & 0.2108    & 94.44\%  & 92.97\%   \\ 
\cline{2-8}
                   & \multirow{3}{*}{\textbf{T=100}} & ResNet15V2       & \multirow{3}{*}{\thead{N=1016 \\
  Test=213}} & 0.0277 & 0.3246    & 98.95\%  & 93.90\%   \\ 
\cline{3-3}\cline{5-8}
                   &                        & VGG-16           &                                    & 0.0648 & \textbf{0.0707}    & 0.9754   & \textbf{98.12\%}   \\ 
\cline{3-3}\cline{5-8}
                   &                        & VGG-19           &                                    & 0.1049 & 0.0821    & 95.01\%  & 97.65\%   \\
\hline
\end{tabular}
\arrayrulecolor{black}
\end{table}
\begin{figure}
\includegraphics[width=\textwidth]{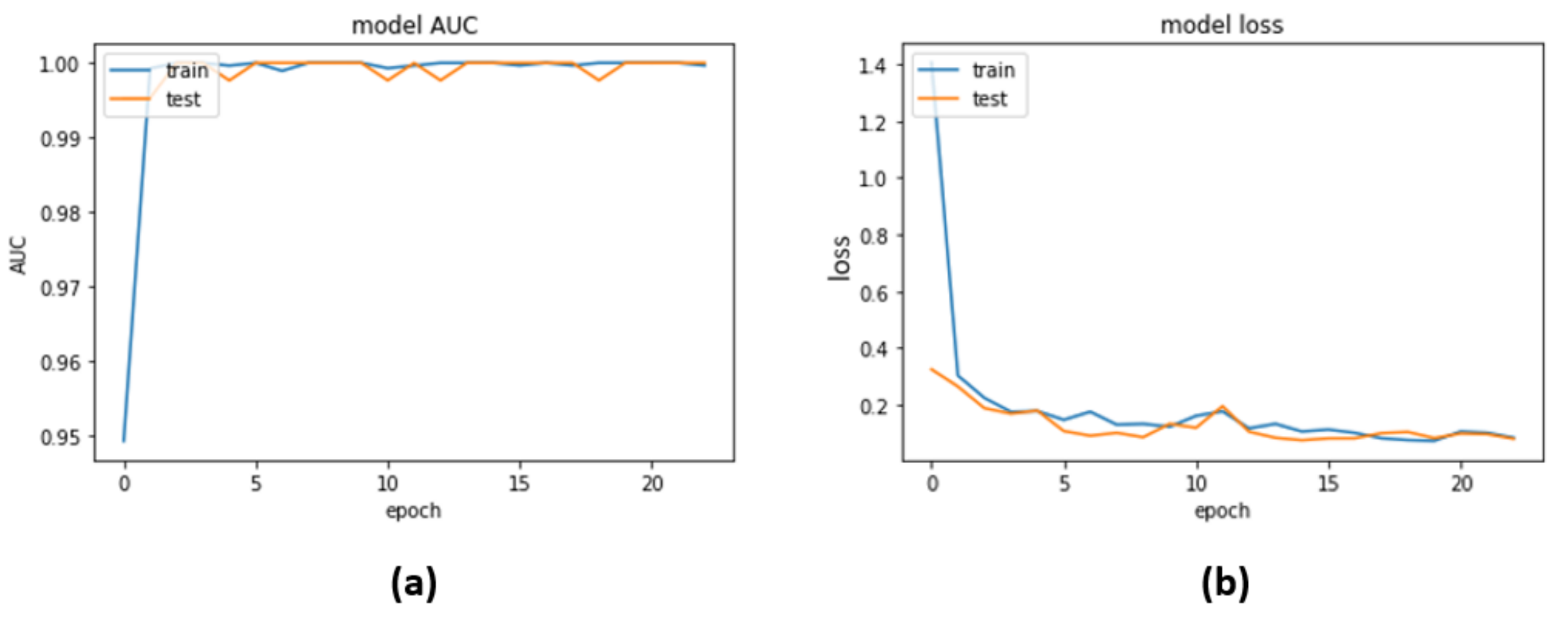}
\caption{Results of (a) train AUC, test AUC, (b) train loss and test loss during the training phase using VGG16 key architecture finetuned on T100 dataset with data augmentation.}

\label{fig5}
\end{figure}

In the first experiment, the performance of the proposed model was lower in the evaluation phase than in training phase. VGG16 gave important results with 91\% of accuracy using T25. However, with the growth of data size the network values did not improve accordingly. On other hand, the performance got higher with both of datasets T50 and T100. There are 5.89\%, 7.93\% and 6.05\% boosts of accuracies with T50 dataset using ResNet152V2, VGG16 and VGG19 models, respectively. In the second experiment, there are remarkable improvements with highest accuracy obtained by VGG16 using T100 dataset. With the presented AUC and loss results in Fig. \ref{fig5} and accuracy value equal to 98.12\%, the network has proven its potential in predicting the rumination behavior. To ensure the reliability and efficiency of the model, we present the sensitivity and precision results in the Table \ref{table:Recallandprecisionofthreenetworks} using T100 dataset.

\begin{table}
\centering
\def\arraystretch{1.3}
\arrayrulecolor{black}
\caption{Recall and precision of three models using the T100.}
\label{table:Recallandprecisionofthreenetworks}
\arrayrulecolor{black}
\begin{tabular}{!{\color{black}\vrule}l!{\color{black}\vrule}l!{\color{black}\vrule}l!{\color{black}\vrule}l!{\color{black}\vrule}l!{\color{black}\vrule}} 
\hline
\multicolumn{2}{!{\color{black}\vrule}l!{\color{black}\vrule}}{~} & \textbf{Recall} & \textbf{Precision} & \textbf{Number of frames}\\ 
\hline
\multirow{2}{*}{\textbf{VGG16}}       & Rumination                         & 99\%   & 97\%      & 110               \\ 
\cline{2-5}
                             & Other                              & 97\%   & 99\%      & 103               \\ 
\hline
\multirow{2}{*}{VGG19}       & Rumination                         & 98\%   & 97\%      & 110               \\ 
\cline{2-5}
                             & Other                              & 97\%   & 98\%      & 103               \\ 
\hline
\multirow{2}{*}{ResNet152V2} & Rumination                         & 98\%   & 91\%      & 110               \\ 
\cline{2-5}
                             & other                              & 89\%   & 98\%      & 103               \\
\hline
\end{tabular}
\arrayrulecolor{black}
\end{table}

\begin{figure}
\includegraphics[width=\textwidth]{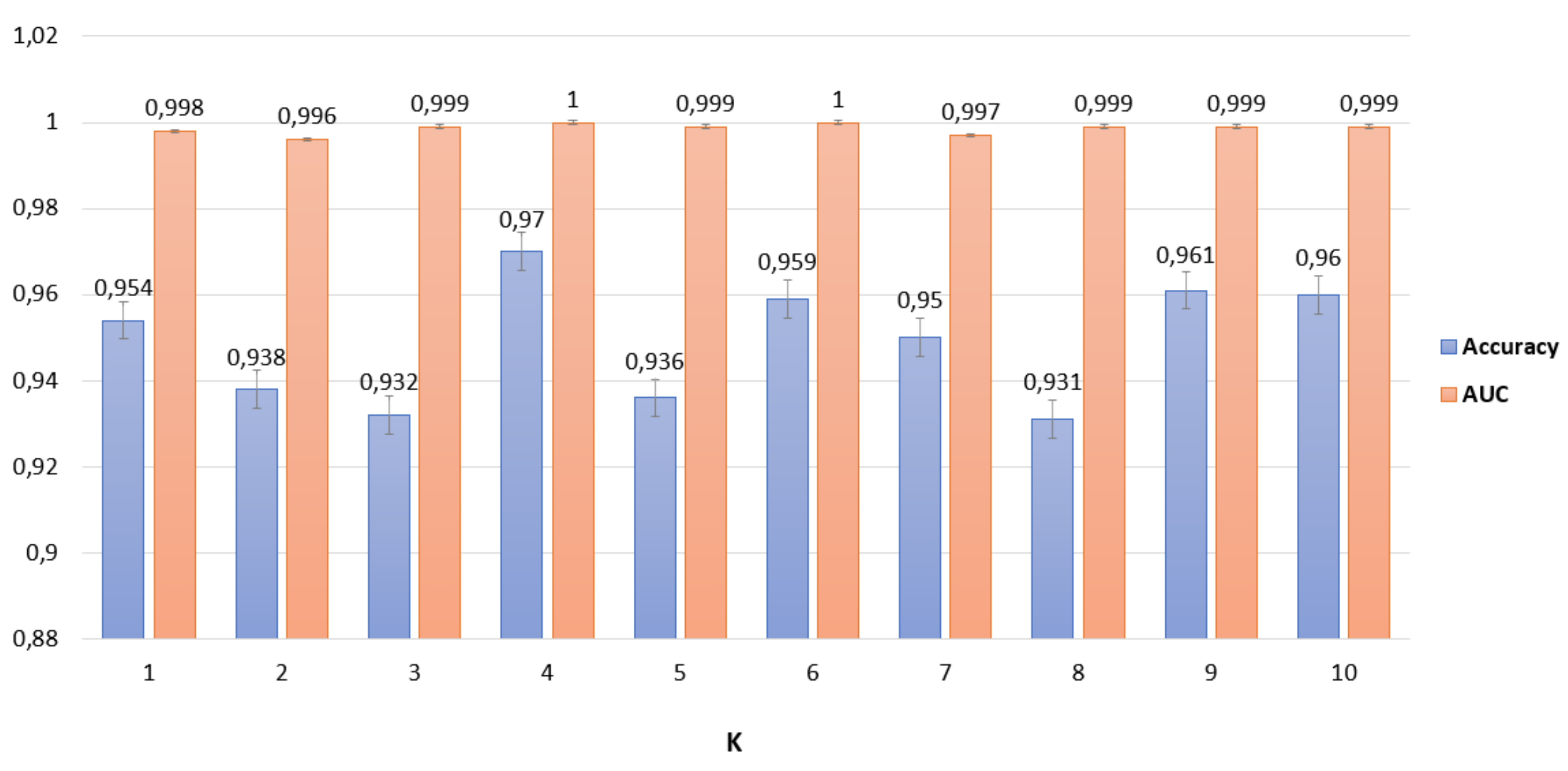}
\caption{Average and STD of the accuracy and AUC metrics using 10 fold cross-validation with VGG 16 as the base network.}

\label{fig5}
\end{figure}

Both of VGG16 and VGG19 achieved higher than 97\% in both precision and recall metrics, which proves the robustness of the network. We notice that VGG16 achieved the best performance by accurately predicting 99\% of rumination behavior. To ensure that the model is performing well with different test sets, we conduct 10 folds cross-validation and present the average, Standard Deviation (STD) values of accuracy and AUC metrics. The results of this procedure are detailed in the Fig. \ref{fig5}, knowing that K is the number of folds.


$\newline$
With these obtained results, the model has proved its potential in predicting and recognizing cow rumination behavior with remarkable highest and lowest average accuracy equal to 93\% and 97\%, respectively. The STD accuracy of the network varies between 2.7\% and 6.9\%. In addition, most of average AUC results are close to 1.00 while the AUC std values are less than 1.2\%, which demonstrate the efficiency and the reliability of our method in recognizing behavior.

\subsection{Comparison}
 
To make the comparison more significant, we compare our proposed method with ResNet50, ResNet152, InceptionV3 \cite{szegedy2016rethinking} and DenseNet121 \cite{huang2017densely} models using T100 generated dataset. The efficiency of the model is done using the accuracy, mean precision and mean recall metrics. The mean precision and recall were calculated using the obtained results during the training stage. The results of the classification are detailed in Table \ref{table:Comparisionofexistingmethods}.

\begin{table}
\centering
\def\arraystretch{1.3}
\caption{Comparison of DenseNet121, InceptionV3, ResNet50 and Resnet152 models with VGG16 architecture using T100 dataset.}
\label{table:Comparisionofexistingmethods}
\arrayrulecolor{black}
\begin{tabular}{|P{2.5cm}|P{2.5cm}|P{2.5cm}|P{2.5cm}|}
\hline
\textbf{Key architecture}  & \textbf{Accuracy} & \textbf{Mean precision} & \textbf{Mean recall}   \\ 
\hline
DenseNet121 & 93\%  & 93.5\%      & 93.5\%         \\ 
\hline
InceptionV3 & 92\%     &   92\%       &    92\%          \\ 
\hline
ResNet50   & 78\%  &  82\%        &  79\%           \\ 
\hline
ResNet152         &  75\%       & 74.5\%    & 74.5\%   \\ 

\hline
 \textbf{VGG16} & \textbf{98\%}    &  \textbf{98\%}      & \textbf{98\%}   \\
\hline
\end{tabular}
\arrayrulecolor{black}
\end{table}

Overall, VGG16 performs favourably against the other architectures. Compared with the presented results, most of models performed less than 98\%. DenseNet121 network achieved 93.5\% in both of mean precision and recall metrics. InceptionV3 gave 92\% of accuracy, recall and precision results. However, both of ResNet50 and ResNet152 performed less than 82\%.

\section{Conclusion}
In this paper, we proposed an effective recognition method with video to monitor and classify cow behavior using deep learning approaches. These technologies proved their potential in complex environments such as farms. They enabled conducting a monitoring method without appealing to these attached and invasive devices. Despite the surrounding inferences (e.g., sunlight and poor lighting) that produced undesirable effects on cow movements such as chewing or swallowing behaviors, we were able to accurately recognize these deep features of rumination behavior using all postures of the dairy cow. Our network basis is simple and easy-to-use based on a standard CNN-based deep learning models. Through an RGB image, the network can recognize long-term dynamics using a compacted representation of a video. The proposed method achieved competitive prediction performance with 98.12\% of accuracy. Future works include the extension of our monitoring method to track rumination time and cows physical activity such as walking and resting.

\section*{Acknowledgment}
This research work is supported by LifeEye LLC. The statements made herein are solely the responsibility of the authors.

\bibliographystyle{IEEEtran}
\bibliography{Simple}

\end{document}